\pdfoutput=1
\documentclass[11pt]{article}

\usepackage[final]{coling}
\usepackage{times}
\usepackage{latexsym}
\usepackage{amsmath}
\usepackage[T1]{fontenc}
\usepackage[utf8]{inputenc}
\usepackage{microtype}
\usepackage{inconsolata}
\usepackage{graphicx}
\usepackage{epsfig}
\usepackage{booktabs} 
\usepackage{epsfig}
\usepackage{amsmath}
\usepackage{amsfonts}
\usepackage{xcolor}
\usepackage{algorithm}
\usepackage[noend]{algpseudocode}
\usepackage{multirow}
\usepackage{mathtools}
\usepackage{subcaption}
\usepackage{wrapfig}
\usepackage{tabulary}
\usepackage{amssymb}
\usepackage{authblk}
\usepackage[toc,page]{appendix}

\newcommand{\subfour}[1]{\vspace*{3mm}{\noindent\bf #1}} 
\newcommand{\subsubfour}[1]{\vspace*{1mm}{\noindent\bf #1}}

\title{Leveraging Taxonomy and LLMs for Improved Multimodal Hierarchical Classification}

\author{\textbf{Shijing Chen}\textsuperscript{1}, \textbf{Mohamed Reda Bouadjenek}\textsuperscript{2}, \textbf{Shoaib Jameel}\textsuperscript{3}, \\\textbf{Usman Naseem}, \textbf{Basem Suleiman}\textsuperscript{1}, \textbf{Flora D. Salim}\textsuperscript{1}, \textbf{Hakim Hacid}\textsuperscript{5}, \textbf{Imran Razzak}\textsuperscript{6,1}  \\

\\

\textsuperscript{1}{University of New South Wales, Australia} \textsuperscript{1}{Deakin University,  Australia} \\
\textsuperscript{3}{University of Southampton, United Kingdom} \textsuperscript{4}{Macquarie University, Sydney, Australia} \\
\textsuperscript{5}{Technology Innovation Institute, UAE} 
\textsuperscript{6}{MBZUAI, Abu Dhabi, UAE}
}

\begin{document}
\maketitle

\begin{abstract}
Multi-level Hierarchical Classification (MLHC) tackles the challenge of categorizing items within a complex, multi-layered class structure. 
However, traditional MLHC classifiers often rely on a backbone model with $n$ independent output layers, which tend to ignore the hierarchical relationships between classes. 
This oversight can lead to inconsistent predictions that violate the underlying taxonomy. 
Leveraging Large Language Models (LLMs), we propose novel taxonomy-embedded transitional LLM-agnostic framework for  multimodality classification. 
The cornerstone of this advancement is the ability of models to enforce consistency across hierarchical levels. 
Our evaluations on the MEP-3M dataset - a Multi-modal E-commerce Product dataset with various hierarchical levels- demonstrated a significant performance improvement compared to conventional LLMs structure. 
\end{abstract}

\section{Introduction}

The increasing complexity of real-world datasets has led to the widespread adoption of multi-level hierarchical structures, making Multi-level Hierarchical Classification (MLHC) a critical tool in modern data analysis \cite{zhang2024teleclass}. 
For instance, large-scale e-commerce platforms like Amazon manage extensive product catalogs through complex taxonomies, which include nested categories, subcategories, and filters (e.g., electronics → laptops → 2-in-1 laptops). 
These taxonomies help users efficiently navigate and refine their search by narrowing down options based on attributes like brand, price range, and features \cite{zhang2024teleclass}. 
In such scenarios, MLHC plays an important role by accurately classifying items within these hierarchies, using the taxonomy to infer relationships and ancestors among categories \cite{boone2022marine}.
This ability to leverage taxonomy-based classification not only enhances user experience but also improves data organization, making MLHC indispensable in sectors dealing with large-scale hierarchical datasets.

\begin{figure}[t]
    \centering
    \begin{subfigure}[b]{0.48\columnwidth}
        \centering
        \includegraphics[width=\textwidth]{ 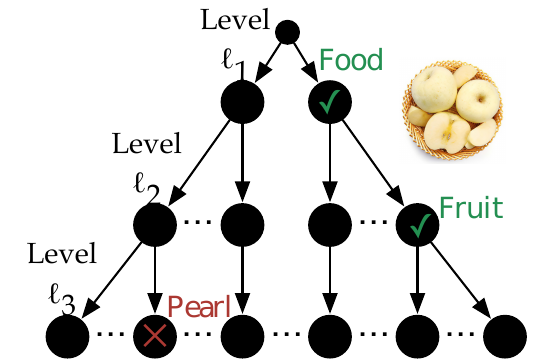} 
        \caption{Taxonomy example.}
        \label{fig:apple}
    \end{subfigure}
    \hfill
    \begin{subfigure}[b]{0.48\columnwidth} 
        \centering
        \includegraphics[width=\textwidth]{ 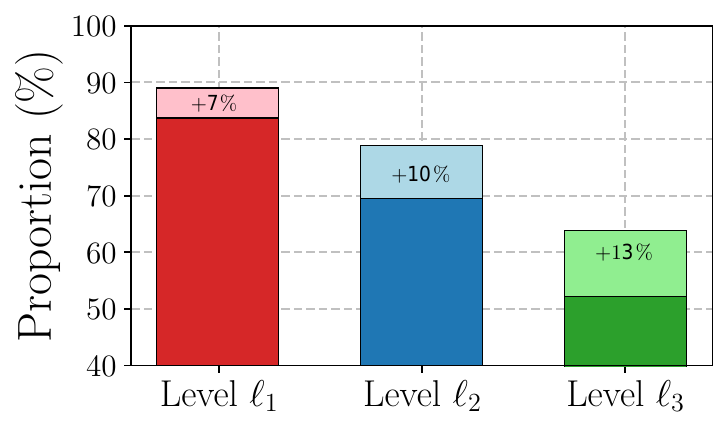} 
        \caption{MEP-3M sampled dataset.}
        \label{fig:MEP}
    \end{subfigure}
    
    \caption{(a) An data point of an ``Apple'' classified by three independent classifiers as a ``Food'' and ``Fruit'' (correct for levels $\ell_1$ and $\ell_2$), but incorrectly classified as a ``Pearl'' at level $\ell_3$. The correct classifications at levels $\ell_1$ and $\ell_2$ could have assisted in identifying the correct class for $\ell_3$. (b) The proportion of correctly classified data entries at each level of the taxonomy for the sampled MEP-3M dataset (shown in dark color), along with the proportion of data entries misclassified at one level but correctly identified at other levels (shown in light color).
    This highlights the potential advantage of using a multi-level hierarchical classifier.}
    \label{fig:combined}
    \vspace{-0.5cm}
\end{figure}

To illustrate and evaluate the benefits of MLHC, we refer to Figure \ref{fig:combined}, which shows
(\ref{fig:apple}) An data entry of an apple classified by three independent classifiers across three hierarchy levels, 
and (\ref{fig:MEP}) the proportion of correctly classified data enties at each level of the taxonomy for the sampled MEP-3M dataset  (shown in dark color), as well as the proportion of data entries that were misclassified at one level but correctly identified at other levels within the taxonomy  (shown in light color). 
Several key insights can be drawn from this analysis: 
(1) First, MLHC facilitates the structure of vast amounts of information using a hierarchical taxonomy, which is particularly useful for capturing the relationships between classes via the ``subclass-of'' concept. 
(2) Second, as demonstrated in Figure~\ref{fig:apple}, providing the final classification layer with information that the data belongs to higher-level categories such as ``Food'' and ``Fruit'' could have enhanced its ability to correctly classify the image at $\ell_3$, or at the very least, ensure consistency by selecting a subclass of ``Fruit''.
(3) Finally, the results depicted in Figure~\ref{fig:MEP} show that 4.08\% of images incorrectly classified at $\ell_1$ were correctly classified at either $\ell_2$ or $\ell_3$, 11.92\% of images misclassified at $\ell_2$ were accurately classified at $\ell_1$ or $\ell_3$, and 16.06\% of images incorrectly classified at $\ell_3$ were correctly classified at $\ell_1$ or $\ell_2$. 
This shows and motivates the potential benefit of an MLHC that embeds the taxonomy structure with a top-down or a bottom-up classification approach.

Several methods have been proposed for MLHC \cite{boone2024mlt}, which can be classified based on how they utilize the hierarchical structure. 
Specifically, we distinguish between three primary approaches: (i) the flat classification approach, where the class hierarchy is completely ignored. 
In this approach, predictions are made solely for the bottom levels, with the assumption that all ancestor classes are implicitly attributed to the instance as well; 
(ii) the local classification approach, which involves training a separate multi-class classifier at each parent node in the hierarchy to distinguish between its child nodes; 
and (iii) the global classification approach \cite{zhang2024teleclass, bettouche2024contextual, liu2024improve}, where a single classifier is responsible for handling the entire class hierarchy.
In this paper, we argue that \textit{flat classifiers}, by ignoring the hierarchical relationships between class levels, often result in inconsistent classifications. 
For instance, as shown in Figure~\ref{fig:apple}, the data entry of an apple is correctly classified as ``Food'' and ``Fruit'', but incorrectly as ``Pearl'' at the leaf node.
Furthermore, we argue that it is impractical to train and maintain $n$ separate networks for \textit{local classification approaches}, which can be redundant and costly in real-world applications. 
As a result, we favor \textit{global classification approaches}, which address the limitations of flat and local methods. 
However, existing methods still face several key challenges: 
(i) they do not inherently embed  the taxonomy structure, 
(ii) they often rely on complex neural network architectures with $n$ independent output layers that do not interact, 
(iii) they frequently produce predictions that are inconsistent with the taxonomy, 
and (iv) they typically operate with a fixed $n$, limiting flexibility and requiring extensive hyperparameter tuning to optimize $n$ for different scenarios.

This paper addresses the aforementioned shortcomings by introducing a novel Taxonomy-based Transitional Classifier (TTC) for MLHC. 
Specifically, we propose an LLM-agnostic output layer that can be used in conjunction with any LLM integrating taxonomy information. 
Our output layer employs a \textit{top-down divide-and-conquer strategy}, attending to the taxonomy relationships at each level of the classifier to ensure predictions remain consistent with the hierarchical structure. 
Focusing on a multimodal dataset, we evaluate the effectiveness of our approach on the MEP-3m dataset~\cite{liu2023mep} and use different LLMs as backbone models. 
Experimental results demonstrate that TTC improves the performance of various backbone LLMs compared to when they are applied as flat classifiers.

\section{Related Work}
Hierarchical classification is a well-established area of research, with a broad spectrum of approaches developed across various domains.
Below, we review the most prominent approaches.

\subsubfour{Flat Classification methods:}
which ignore the hierarchical structure, typically predicting only classes at the leaf nodes and considering that all its ancestor classes are also implicitly assigned to that instance.
Although these methods are simple and computationally efficient, they often fail to utilize the inherent relationships between classes, resulting in suboptimal performance, particularly in complex taxonomies~\cite{silla2011survey, valentini2010true}. 
While commonly used as baselines in empirical studies, their lack of hierarchical awareness limits their effectiveness in domains where class relationships are crucial.

\subsubfour{Local Classifier Approaches:} have been widely adopted to address the limitations of flat classification by training classifiers at different levels of the hierarchy. 
These approaches can be further divided into three types: Local Classifier per Node (LCN), Local Classifier per Parent Node (LCPN), and Local Classifier per Level (LCL). 
The LCN approach, proposed by Koller and Sahami \cite{koller1997hierarchically}, is perhaps the most common, where a classifier is trained for each node in the hierarchy. 
However, it is prone to inconsistencies in predictions across levels \cite{silla2011survey,dumais2000hierarchical}. 
The LCPN approach trains classifiers for each parent node to distinguish among its children, which can reduce inconsistencies but may still propagate errors down the hierarchy \cite{secker2007experimental}. 
The LCL approach, though less commonly used, involves training classifiers at each level of the hierarchy, but it can suffer from the challenge of discriminating among a large number of classes at deeper levels~\cite{de2009tutorial,costa2007comparing}.

\subsubfour{Global Classifier Approaches:} in contrast, treat the entire hierarchy as a single unit during training. 
These approaches integrate hierarchical information into the learning process, ensuring consistency across levels but often at the cost of increased computational complexity and reduced modularity \cite{vens2008decision}. 
Notable examples include the Clus-HMC algorithm, which uses predictive clustering trees to handle the hierarchical structure \cite{kiritchenko2005functional, vens2008decision}. 
Global classifiers are advantageous in that they avoid the error propagation issues inherent in local approaches, but they require significant computational resources and are often specific to the underlying flat classifier being adapted \cite{silla2011survey}.


\subsubfour{Graph Neural Networks (GNNs):} have become a central tool in hierarchical classification, particularly for handling complex dependencies between labels. 
These networks model the entire hierarchy as a graph, where nodes represent labels and edges represent hierarchical relationships. 
Recent works have shown that GNNs are particularly effective in capturing both horizontal and vertical dependencies within the hierarchy, leading to improvements in classification performance across multiple levels such as models like Hierarchy-Aware Graph Models (HiAGM) \cite{liu2023enhancing}.

\subsubfour{Specialized loss functions:} have emerged as a robust method for ensuring consistency in hierarchical multi-label classification. 
This approach is designed to handle the intricacies of pre-defined class hierarchies by incorporating a max constraint loss (MCLoss) that enforces hierarchical dependencies during training. 
The Coherent Hierarchical Multi-Label Classification Networks (C-HMCNN) \cite{giunchiglia2020coherent} are proposed with such MCLoss to ensure that the predictions across the hierarchy remain coherent, meaning that a child node can only be activated if its parent node is also activated. 
This method effectively maintains logical consistency across hierarchical levels, significantly improving classification accuracy where adherence to the hierarchy is critical.

\subsubfour{Advances and Challenges: }
Recent advancements in hierarchical classification have focused on integrating deep learning techniques and graph-based approaches, particularly for tasks involving multi-level taxonomies, such as document categorization and other NLP tasks. 
Despite progress, challenges remain in scaling models to handle large, complex hierarchies consistently \cite{boone2022mask}. 
The Taxonomy-based Transitional Classifier (TTC) proposed in this paper addresses these issues by embedding hierarchical information directly into the classification process and leveraging LLMs for multi-modal data. 
As a model-agnostic layer, TTC enhances flexibility and accuracy across various backbone models, offering a more consistent solution for complex hierarchies.

\section{Taxonomy-based Transitional Classifier}
This section formally defines the MLHC problem and introduces our proposed Taxonomy-based Transitional Classifier, designed to enforce hierarchical consistency across classification levels.

\subsection{Notation and problem definition}

\subsubfour{Classification:} 
Most classification problems in the literature focus on flat classification, where each instance is assigned to a single class from a finite set of independent, non-hierarchical classes.
Formally, given a dataset $\mathcal{D} = \{(\textbf{x}^{(1)},y^{(1)}), (\textbf{x}^{(2)},y^{(2)}),$ $\cdots, (\textbf{x}^{(m)},y^{(m)})\}$ with $m$ instances, where each $\textbf{x}^{(i)}\in \mathbb{X} \subseteq \mathbb{R}^n$ is an $n$-dimensional input feature vector of the instance $i$ and $y^{(i)}\in \mathcal{Y}=\{y_1,y_2,\cdots,y_k\}$ represents its class, 
a classification algorithm must learn a mapping function $f:\mathbb{X} \rightarrow \mathcal{Y}$, which assigns to each feature vector $\textbf{x}^{(i)}$  its correct class $y^{(i)}$.

\subsubfour{Multimodal classification:} 
It extends the flat classification paradigm by incorporating multiple data modalities, such as text, images, or audio, into the classification process. 
Formally, given a multimodal dataset $\mathcal{D} = \{((\textbf{x}_1^{(i)}, \textbf{x}_2^{(i)}, \cdots, \textbf{x}_p^{(i)}), y^{(i)}) \mid i = 1, 2, \cdots, m\}$, where $\textbf{x}_j^{(i)} \in \mathbb{X}_j \subseteq \mathbb{R}^{n_j}$ represents the feature vector of modality $j$ for instance $i$, and $y^{(i)} \in \mathcal{Y}$ represents the class, a multimodal classification algorithm learns a mapping function $f: (\mathbb{X}_1 \times \mathbb{X}_2 \times \cdots \times \mathbb{X}_p) \rightarrow \mathcal{Y}$. This function assigns the correct class label $y^{(i)}$ by leveraging information from all available modalities to improve prediction accuracy and robustness.

\subsubfour{Hierarchical classification:} 
In contrast to \textit{flat classification} in which classes are considered unrelated, in a hierarchical classification problem classes are organized in a taxonomy. 
The taxonomy is often organized as a tree, where classes have a single parent each, or a directed acyclic graph (DAG), where classes can have multiple parents.
Given a set of classes $\mathcal{Y}$, Wu et al.~\cite{wu2005learning} defined a taxonomy as a pair $(\mathcal{Y},  \prec)$, where $\prec$ is the \textit{``subclass-of''} relationship with the following properties~\cite{wu2005learning,silla2011survey}: 
(i) asymmetry ($\forall y_i, y_j \in \mathcal{Y}, if y_i \prec y_j$ then $y_j \nprec y_i$), 
(ii) anti-reflexivity ($\forall y_i \in \mathcal{Y}, y_i \nprec y_i)$, 
and (iii) transitivity ($\forall y_i, y_j, y_k \in \mathcal{Y}, y_i \prec y_j$ and $y_j \prec y_k$ implies $y_i \prec y_k$).

In this paper, we consider only \textit{tree} taxonomies, which are organized with a hierarchy structure of $n$ levels $\ell_i$, such that $\ell_i \subset \mathcal{Y}$, $\ell_1 \cup \ell_2  \cdots \cup \ell_n = \mathcal{Y}$, $\forall y_j \in \ell_1,  y_i \prec \emptyset $, and $\forall y_j \in \ell_{i+1}, \exists! y_k \in \ell_{i}$   s.t. $ y_j \prec y_k$ for $i\ge 1$
(see Figure~\ref{fig:apple} for a three-level taxonomy).
Finally, we encode the relationship between two successive levels $\ell_i$ and  $\ell_{i+1}$ in a taxonomy  using an $|\ell_i| \times |\ell_{i+1}|$  matrix $M^{[\ell_i,\ell_{i+1}]}$, where the binary value $M^{[\ell_i,\ell_{i+1}]}_{y_k,y_j} \in \{0 (y_j \nprec y_k), 1 (y_j \prec y_k) \}$, with $y_k \in \ell_i$ and $y_j \in \ell_{i+1}$.

\subsubfour{Problem definition:} 
The multimodal multi-level hierarchical classification problem addressed in this paper is defined as the task of learning a mapping function
$f:(\mathbb{X}_1 \times \mathbb{X}_2 \times \cdots \times \mathbb{X}_p) \rightarrow \mathcal{Y}$, which assigns to each instance--represented by a combination of feature vectors from $p$ different modalities--a prediction vector $\textbf{y}^{(i)} = \{y^{[\ell_1]}, y^{[\ell_2]},\cdots, y^{[\ell_n]}\}$. 
Here, $y^{[\ell_i]} \in \ell_i$ represents the class assigned by the function $f$ at each hierarchical level $\ell_i$, ensuring accurate predictions across all taxonomy levels.

\subsection{TTC Model Description}
As mentioned earlier, our proposed TTC addresses the limitations of existing methods, which often result in inconsistent predictions, by enforcing consistency throughout the prediction process.
Leveraging the detailed taxonomic information at each hierarchical level, the TTC layer guides its predictions by restricting them to labels that are appropriate for the corresponding level in the hierarchy. 
This approach avoids inconsistent classifications that span unrelated categories. 
By integrating the hierarchical structure directly, the TTC layer promotes consistency and aims to enhance the logical soundness of predictions in a multimodal context, potentially achieving better overall accuracy than conventional LLMs.

\begin{figure}[t]
    \centering
    \includegraphics[width=0.99\columnwidth]{ 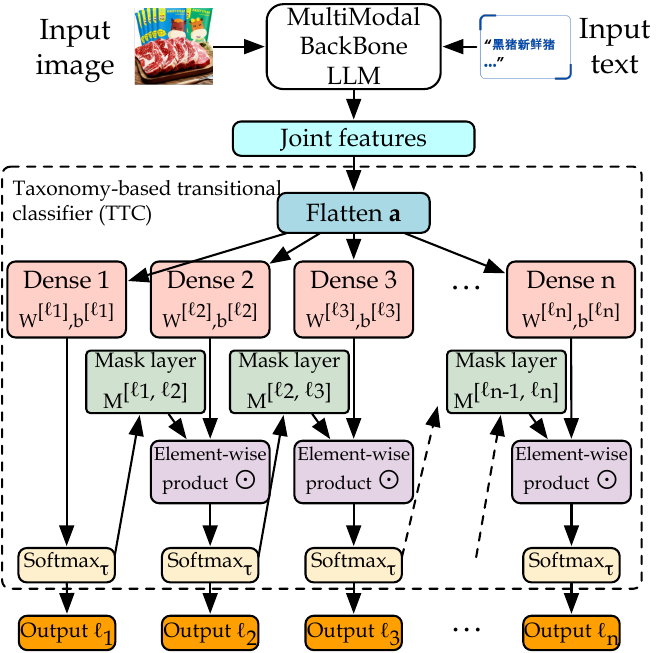}
    \caption{Architecture diagram of the taxonomy-based transitional classifier. The transitional matrix \(M^{[\ell_n, \ell_{n+1}]}\) multiplies the output from the corresponding classifier to obtain an attention score, which is then applied to the output of the next-level classifier. This ensures that the information from the upper level and subclass predictions is integrated into the output, increasing the likelihood of maintaining consistency.}
    \label{fig:taxonomy}
    \vspace{-0.5cm}
\end{figure}

Figure \ref{fig:taxonomy} illustrates the architecture of the proposed TTC layer, an LLM-agnostic component designed to leverage the taxonomy and ensure that predictions adhere to the hierarchical structure of the data.
Several independent classifiers are used to predict the categories on different levels in the same way as local approaches. 
However, to maintain consistency, the relation information of upper levels is incorporated into the next level in the same way as attention is. 
The output probabilities from the upper level are multiplied by a \textit{transition} matrix, where each entry represents the relationship between classes at successive levels in the taxonomy (i.e., 1 if the class in the column is a ``subclass of'' the class in the row, and 0 otherwise).
The product can be considered as the attention score that incorporates the hierarchical information as well as the relation between classes and can be applied to the output probability for the next level. 
The prediction of the classifiers can be formulated as $\textbf{z}^{[\ell_i]} = W^{[\ell_i]} \cdot \textbf{a} + b^{[\ell_i]}$, where $\textbf{a}$ is the joint output latent feature of backbone multimodal LLMs, and $W^{[\ell_i]}, b^{[\ell_i]}$ are learnable parameters that trained on the trainset regarding each $\ell_i$ of the hierarchies. 
The prediction of the first classifier is obtained by applying a temperature-scaled \textit{softmax} normalization, as $\hat{\textbf{y}}^{[\ell_1]} = \text{softmax}(\textbf{z}^{[\ell_1]})$. 
For each subsequent level, we compute an attention score to incorporate relational information into the predictions, ensuring consistency across levels (i.e., $\hat{y}^{[\ell_{i+1}]} \prec \hat{y}^{[\ell_i]}$). This is achieved by injecting hierarchical relations as follows:
\begin{equation} \label{equation-attention}
\textbf{m}^{[\ell_{i+1}]} = \hat{\textbf{y}}^{[\ell_i]} \times M^{[\ell_i, \ell_{i+1}]}
\end{equation}

\noindent where \(M^{[\ell_i, \ell_{i+1}]}\) is our \( |\ell_i| \times |\ell_{i+1}| \) transitional matrix which encodes the relationship between two successive levels \( \ell_i \) and \( \ell_{i+1} \) in a taxonomy (i.e., the binary value \( M^{[\ell_i, \ell_{i+1}]}_{y_k,y_j} \in \{ 0 \ (\text{if } y_j \not\prec y_k), 1 \ (\text{if } y_j \prec y_k) \} \), with \( y_k \in \ell_i \) and \( y_j \in \ell_{i+1} \)). 
Referring to the example illustrated in Figure \ref{fig:apple}, consider the \(\ell_2\) labels, which include \textit{Jewel} and \textit{Fruit}, and the \(\ell_3\) labels, comprising \textit{K gold}, \textit{Pearl}, \textit{Apple}, and \textit{Pear}. 
The corresponding transition matrix \(M^{[\ell_2, \ell_3]}\) is:

\[
    M^{[\ell_2, \ell_3]} = 
    \begin{pmatrix}
        1 & 1 & 0 & 0 \\
        0 & 0 & 1 & 1
    \end{pmatrix}
\]

\noindent in which the first row corresponds to the  \(\ell_2\)  class \textit{Jewel}, where a value of 1 indicates that the  \(\ell_3\)   class (e.g., \textit{K gold} or \textit{Pearl}) is a subclass of \textit{Jewel}, and a value of 0 indicates no such relationship. 
Similarly, the second row refers to the \(\ell_2\) class \textit{Fruit}, where the values reflect whether the \(\ell_3\) classes are subclasses of \textit{Fruit}. 
In this manner, the hierarchical structure of the taxonomy is fully encapsulated within the transition matrix.

Each attention score is applied using an element-wise product on the probability output of each classifier from a lower level  as:
\begin{equation} \label{equation-predictions}
    \hat{\textbf{y}}^{[\ell_{i+1}]} = \textit{softmax}_{\tau} (\textbf{z}^{[\ell_{i+1}]} \circ \textbf{m}^{[\ell_{i+1}]})
\end{equation}

Attention scores and classifications in Equations~\ref{equation-attention} and \ref{equation-predictions}, respectively, are processed sequentially for all hierarchical levels. The loss function is also adjusted as follows:

\begin{equation}
\frac{1}{m} \sum_{j=1}^{m} \sum_{i=1}^{n} \left[ \pi^{[\ell_i]} \cdot \mathcal{L}(y^{(j)}_{[\ell_i]}, \hat{y}^{(j)}_{[\ell_i]}) \right]
\end{equation}

where $\mathcal{L} (\bullet,\bullet )$ denotes the cross-entropy function and $\pi^{[\ell_i]}$ are a set of importance factors that can be tuned to changing the weight of losses for different \(\ell_i\). 

Continuing with the example provided earlier, given the \textit{transition} matrix \(M^{[\ell_2, \ell_3]}\), and assuming the probability output from the \(\ell_2\) classifier is \(\hat{\textbf{y}}^{[\ell_2]} = \{0.9, 0.1\}\), the attention scores are calculated as: 
\(
\textbf{m}^{[\ell_3]} = \hat{\textbf{y}}^{[\ell_2]} \cdot M^{[\ell_2, \ell_3]} = \{0.9, 0.9, 0.1, 0.1\}
\).
Assuming the output from \(\ell_3\) is \(\textbf{z}^{[\ell_3]} = \{-0.2, 0.5, 1.3, 0.3\}\), applying the attention scores \(\textbf{m}^{[\ell_3]}\) and a softmax function to normalize the result gives the prediction probability output:
\(
\hat{\textbf{y}}^{[\ell_3]} = \{0.182, 0.342, 0.249, 0.225\}.
\)

Compared to a flat classifier for \(\ell_3\) which would have applied directly \textit{softmax} to \(\textbf{z}^{[\ell_3]}\),  TTC's prediction produces more consistency with upper-level prediction. 
Additionally, from a taxonomic perspective, \textit{tree-like} hierarchical classification leverages general-to-specific relationships, where general categories have better data separability. This indicates that they possess wider margins in their decision boundaries, making it easier for classifiers to distinguish them. As a result, general classes at higher levels contribute to higher classification accuracy at the top \cite{cortes1995support}. By enforcing the consistency across hierarchical levels, the LLM is further guided to make more accurate predictions at deeper, more specific levels with greater granularity.

\section{Experiments and Results}
\begin{figure}[t]
    \centering
    \includegraphics[width=0.9985\columnwidth]{ 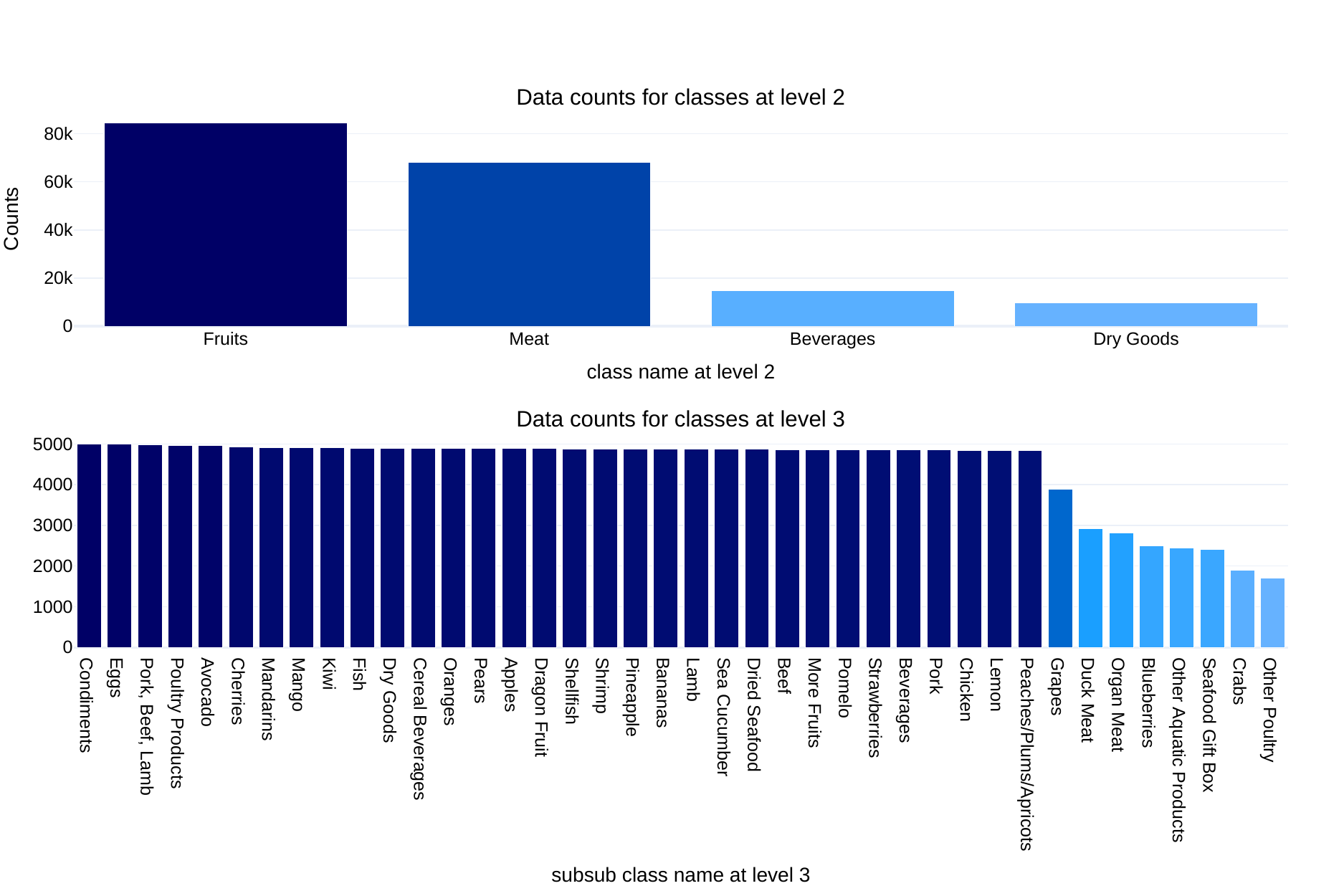}
    \caption{The distribution of data across all classes at $\ell_2$ and $\ell_3$}
    \label{fig: labels}
\end{figure}

This section evaluates the performance of a TTC in MLHC tasks using the MEP-3 dataset, a large-scale multimodal e-commerce product dataset containing over 3 million entries. Due to computational constraints and a significant portion of entries lacking third-level hierarchical labels, we focused on the third-largest food subset, chosen for its diversity in product types. 
To maintain consistency in the experiments, we excluded entries that do not include third-level hierarchies. 
As a result, the final dataset used for these experiments consisted of 177,195 data points. 
Figure \ref{fig: labels} illustrates the distribution of data across all classes at both hierarchies.

\begin{table}[h]
\centering
\caption{Hyperparameters used in the experiments.}
\label{tab:hyperparameters}
\resizebox{\columnwidth}{!}{%
\begin{tabular}{lc}
\hline \hline
\textbf{Hyperparameter}       & \textbf{Value} \\ \hline\hline
Learning Rate (Fine-Tuning)   & 1e-5           \\ \hline
Learning Rate (TTC Training)  & 1e-4           \\ \hline
Batch Size                    & 32             \\ \hline
Number of Epochs              &  (3 tolerance on train acc) \\ \hline
LoRA Rank                     & 8              \\ \hline
LoRA alpha                     & 16              \\ \hline
LoRA dropout                     & 0.1              \\ \hline
Weight Decay                  & 0.01           \\ \hline
Dropout Rate                  & 0.1            \\ \hline
Optimizer                     & AdamW          \\ \hline \hline
\end{tabular}%
}

\end{table}

\subsection{Experimental Details}
\subfour{Experimental Setting: } The preprocessing steps were as follows: \textit{Textual Data}: Product descriptions were tokenized using Byte Pair Encoding (BPE) or Wordpiece tokenization, which are commonly used methods for handling both Chinese and multilingual text. Stop words were also removed to reduce noise, improving the model's focus on relevant content. \textit{Image Data}: The majority of images in the dataset were already $220\times220$ pixels, but a minority were smaller, with sizes like $64\times50$, $75\times75$, $60\times60$, and $54\times54$. To ensure uniformity across the dataset, all images were resized to 220x220 pixels. This resizing helped maintain consistent input dimensions for the model. Additionally, pixel values were normalized to fall within a common range to improve model stability and convergence during training. \textit{Hierarchical Labels}: The dataset's hierarchical structure consists of three levels: 1 top-level class, 4 second-level sub-classes, and 40 third-level sub-sub-classes. Labels were encoded to ensure that each product was accurately categorized across the relevant levels of hierarchy.

An 80/20 split was applied to this food subset, with 80\% of the data (141,756 inputs) used for training.
 We split the whole process into two stages to speed up the training procedure: \textit{Fine-tune LLMs}: We applied Low-Rank Adaptation (LoRA) \cite{hu2021lora} to the LLMs and fine-tuned them on the training set to improve the models' representation capabilities. Since bottom-level labels offer greater granularity, the model learned finer distinctions between similar classes within broader categories. To achieve this, we sampled 1,000 data entries from each \(\ell_2\) label (40,000 in total). This balanced sampling provided sufficient data for the model to learn detailed and specific features for each class, which is essential for fine-grained classification. By ensuring that each \(\ell_2\) class had an equal number of samples, we avoided overrepresentation of certain classes, leading to more stable and reliable performance across all categories. We utilized the Parameter-Efficient Fine-Tuning (PEFT) library \cite{peft} to efficiently apply LoRA to the candidate backbone LLMs. \textit{Train TTC}: Aligned with the hierarchical levels of the dataset we used, the TTC also consists of 2 levels of classifiers to predict both levels of categories. After fine-tuning, the backbone models are integrated with TTC with frozen parameters to train the classifier further on the training set. 20\% (35,439 inputs) of the subset is used for testing. The input of the dataset, which contains \((v, t)\), where \(v\) refers to the representation from the image and \(t\) refers to the textual representation, is fed into the backbone Multimodal LLMs to generate a joint implicit feature \(x\) for the taxonomy-based transitional classifier to make predictions on each level of hierarchy.

\begin{figure*}[t]
    \centering
    \includegraphics[width=1\textwidth]{ 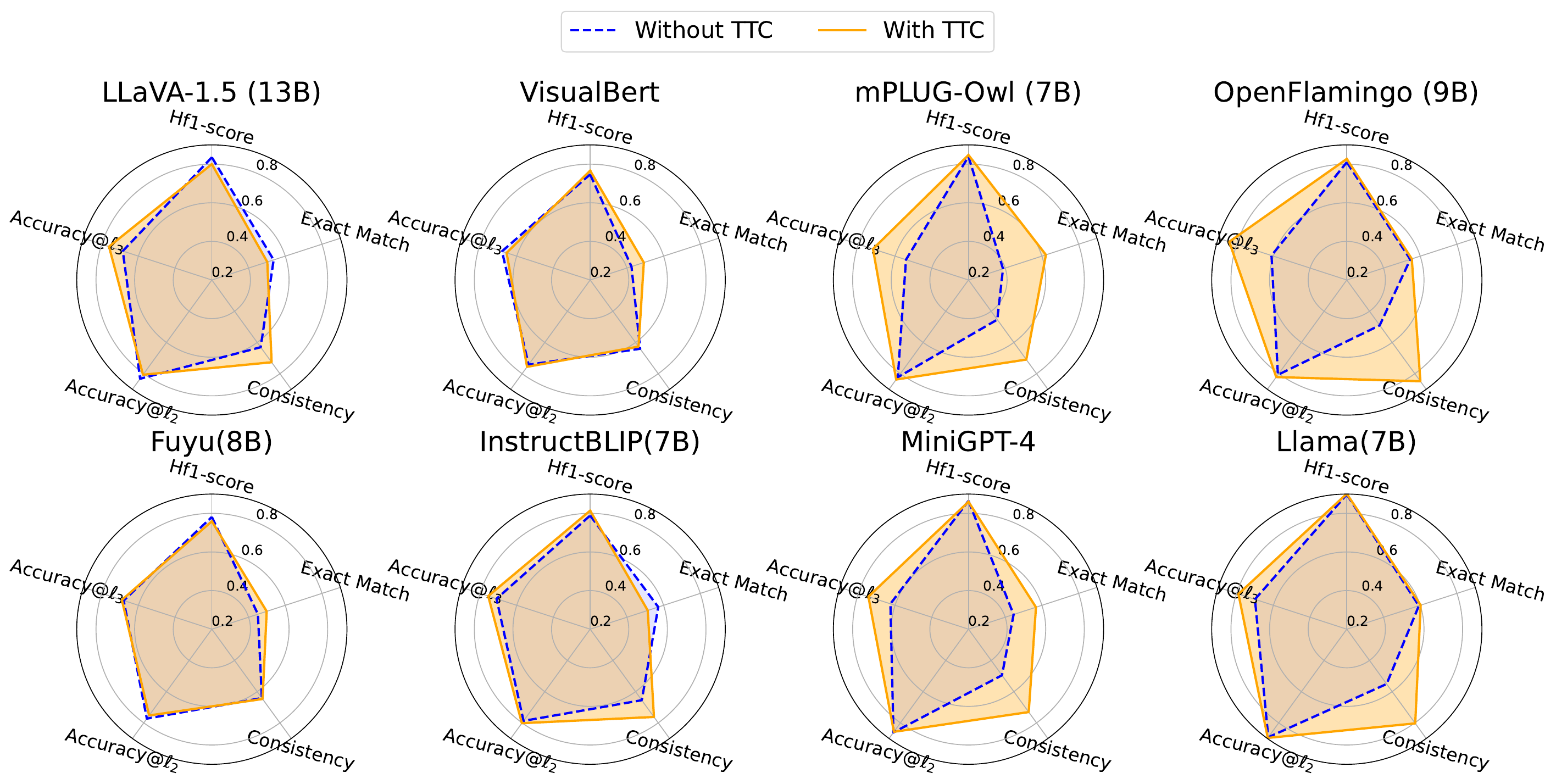}
    \caption{Spider diagram for all LLM backbones regarding 5 different metrics.}
    \label{fig: spider}
\end{figure*}

\begin{table*}[t]
    \centering
    \caption{Performance of Multimodal Large Language Models with and without TTC on the newest MEP-3 dataset.}
    \label{tab:results}
    \vspace{-0.3cm}
\small
    \begin{tabular}{lccccc}
        \hline \hline
        \textbf{Food Subset} & \textbf{HF1-Score} & \textbf{Exact Match} & \textbf{Consistency} & \textbf{Accuracy@$\ell_2$} & \textbf{Accuracy@$\ell_3$} \\
        \hline \hline
        LLAVA-1.5 (13B) & \textbf{0.8356} & \textbf{0.5367} & 0.6305 & 0.8324 & 0.6853 \\
        LLAVA-1.5 (13B) + TTC & 0.8003 & 0.5021 & \textbf{0.7268} & \textbf{0.8067} & \textbf{0.7601} \\ \hline\hline
        VisualBert & 0.7472 & 0.4251 & 0.6405 & 0.7423 & 0.6804 \\
        VisualBert + TTC & \textbf{0.7663 }& \textbf{0.4923} & \textbf{0.6263} & \textbf{0.7562} & \textbf{0.6554} \\ \hline \hline
        mPLUG-Owl (7B) & 0.8396 & 0.3881 & 0.4538 & 0.8229 & 0.5404 \\
        mPLUG-Owl (7B) + TTC & \textbf{0.8475} & \textbf{0.6218} & \textbf{0.7096} & \textbf{0.8373} &\textbf{ 0.7203} \\ \hline \hline
        OpenFlamingo (9B) & 0.8102 & 0.5457 & 0.4903 & 0.8071 & 0.6102 \\
        OpenFlamingo (9B) + TTC & \textbf{0.8273} & \textbf{0.5541} & \textbf{0.8485} & \textbf{0.8218} &\textbf{ 0.8404} \\ \hline \hline
        Fuyu (8B) & \textbf{0.7815} & 0.4510 & 0.6409 & \textbf{0.7728} & 0.6803 \\
        Fuyu (8B) + TTC & 0.7591 & \textbf{0.4988 }& \textbf{0.6471} & 0.7526 & \textbf{0.6902 }\\ \hline \hline
        InstructBLIP (7B) & 0.7888 & \textbf{0.5710} & 0.6549 & 0.7874 & 0.7101 \\
        InstructBLIP (7B) + TTC & \textbf{0.8140} & 0.5137 & \textbf{0.7622 }& \textbf{0.8021} & \textbf{0.7554} \\ \hline \hline
        MiniGPT-4 & 0.8649 & 0.4458 & 0.4952 & 0.8615 & 0.6253 \\
        MiniGPT-4 + TTC & \textbf{0.8652} & \textbf{0.5677} & \textbf{0.7309} & \textbf{0.8573} & \textbf{0.7453} \\ \hline \hline
        Llama (7B) & 0.8979 & 0.5926 & 0.5510 & 0.8921 & 0.7004 \\
        Llama (7B) + TTC & \textbf{0.9016} & \textbf{0.6024} & \textbf{0.8033} & \textbf{0.8964} & \textbf{0.7903} \\
        \hline \hline
    \end{tabular}
    \vspace{-0.3cm}
\end{table*}

\subfour{Backbone Multimodal LLMs:} 
For backbone models, we have adopted different LLMs including: LLAVA-1.5 (13B) \cite{liu2024visual}, VisualBert \cite{li2019visualbert}, mPLUG-Owl (7B) \cite{ye2023mplug}, OpenFlamingo(9B) \cite{awadalla2023openflamingo}, Fuyu(8B) \cite{Bavishi2023Fuyu}, InstructBLIP(7B) \cite{dai2024instructblip}, MiniGPT-4 \cite{zhu2023minigpt}, and Llama (7B) \cite{touvron2023llama}. 
Notably, for Llama, we manually integrated ResNet as the visual encoder to enable multimodal functionality. 
ResNet extracts visual features from images, which are then projected to align with Llama’s embedding dimensions and fused with textual inputs. 
This integration allows Llama to process and generate responses based on both visual and textual information. 
Each model is assessed with and without the proposed method, denoted by (TTC). 
The hyperparameters for the experiments are set as Table \ref{tab:hyperparameters}.

\subfour{Evaluation Metrics:} For evaluating the MLHC task, we have adopted the Hierarchical F1-Score (HF1- score)~\cite{kosmopoulos2015evaluation}, which assesses model performance in predicting classes across different hierarchy levels. Similar to the F1-score, the HF1-Score is defined as: 

$$\text{HF1- Score} = \frac{2 \cdot (\text{H-Precision} \cdot \text{H-Recall})}{\text{H-Precision} + \text{H-Recall}}$$. 

\noindent where H-Recall and H-Precision are analogous to Recall and Precision but evaluate the proportion of correctly predicted classes among all actual/predicted classes. In addition to the HF1-Score, we also use consistency and Exact Match as evaluation metrics. Consistency ensures that predicted labels adhere to hierarchical structures, meaning predictions across all levels remain within the same hierarchy. Exact Match is a stricter criterion requiring that predictions not only stay within hierarchy but also exactly match true labels at all levels.

\begin{figure*}[t]
    \centering
   \includegraphics[width=0.9985\textwidth]{ 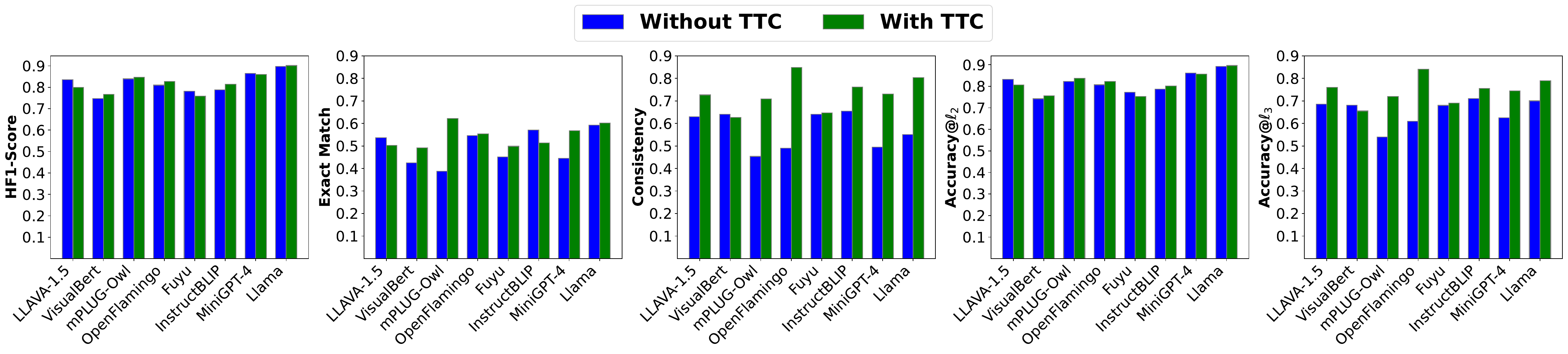}
    \caption{Group bar chart for detailed results for all LLM backbones on all 5 metrics.}
    \label{fig:bar}
    \vspace{-0.5cm}
\end{figure*}

\subsection{Experimental Results}


We present the results of applying the proposed taxonomy-based transitional classifier (TTC) to various large multimodal LLMs for a comparative analysis. Figure \ref{fig: spider} provides an overall comparison of the different LLMs with and without the integration of the model-agnostic TTC layer (see Table \ref{tab:results} for details, and refer to Appendix \ref{sec:appendix} for further information). Across the board, integrating the hierarchical layer shows a clear improvement in performance for most models, confirming the efficacy of the TTC approach. In particular, while \textbf{HF1-Score} remains relatively high across models—indicating a strong ability to capture hierarchical relationships—there is a slight decrease in this metric for some models when TTC is introduced. This suggests that TTC's emphasis on enforcing consistency between layers can result in a trade-off with general performance. However, the hierarchical layer consistently leads to improvements in \textbf{Consistency}, \textbf{Exact Match}, and \textbf{ Accuracy} at $\ell_3$, highlighting its strength in producing more coherent and fine-grained predictions. These enhancements underline the effectiveness of TTC in addressing complex hierarchical classification tasks, ensuring predictions align better with structured taxonomy.

For detailed comparison, Figure \ref{fig:bar} further highlights the impact of TTC by illustrating how it significantly boosts key metrics such as \textbf{Consistency} and \textbf{Exact Match} across models. For example, while applying TTC to LLAVA-1.5 (13B) resulted in a slight reduction in \textbf{HF1-Score} (from 0.8356 to 0.8003), it led to a substantial improvement in \textbf{Consistency}, increasing from 0.6305 to 0.7268, indicating more coherent predictions aligned with the hierarchical structure. Similarly, mPLUG-Owl (7B) saw a remarkable improvement in \textbf{Exact Match} (from 0.3881 to 0.6218) and \textbf{Consistency} (from 0.4538 to 0.7096), demonstrating TTC's ability to enhance alignment with taxonomical classifications. OpenFlamingo (9B), which experienced improvements across all metrics, particularly in \textbf{Consistency} (from 0.4903 to 0.8485) and \textbf{$\ell_3$ Accuracy} (from 0.6102 to 0.8404), further reinforces the effectiveness of TTC in producing more precise and reliable predictions. These results collectively showcase TTC's capacity to significantly improve LLMs' ability to handle hierarchical classification tasks, leading to more accurate and consistent model outputs, especially for tasks that require deeper, fine-grained distinctions.

\begin{figure}[t]
    \centering
    \includegraphics[width=0.9979285\columnwidth]{ 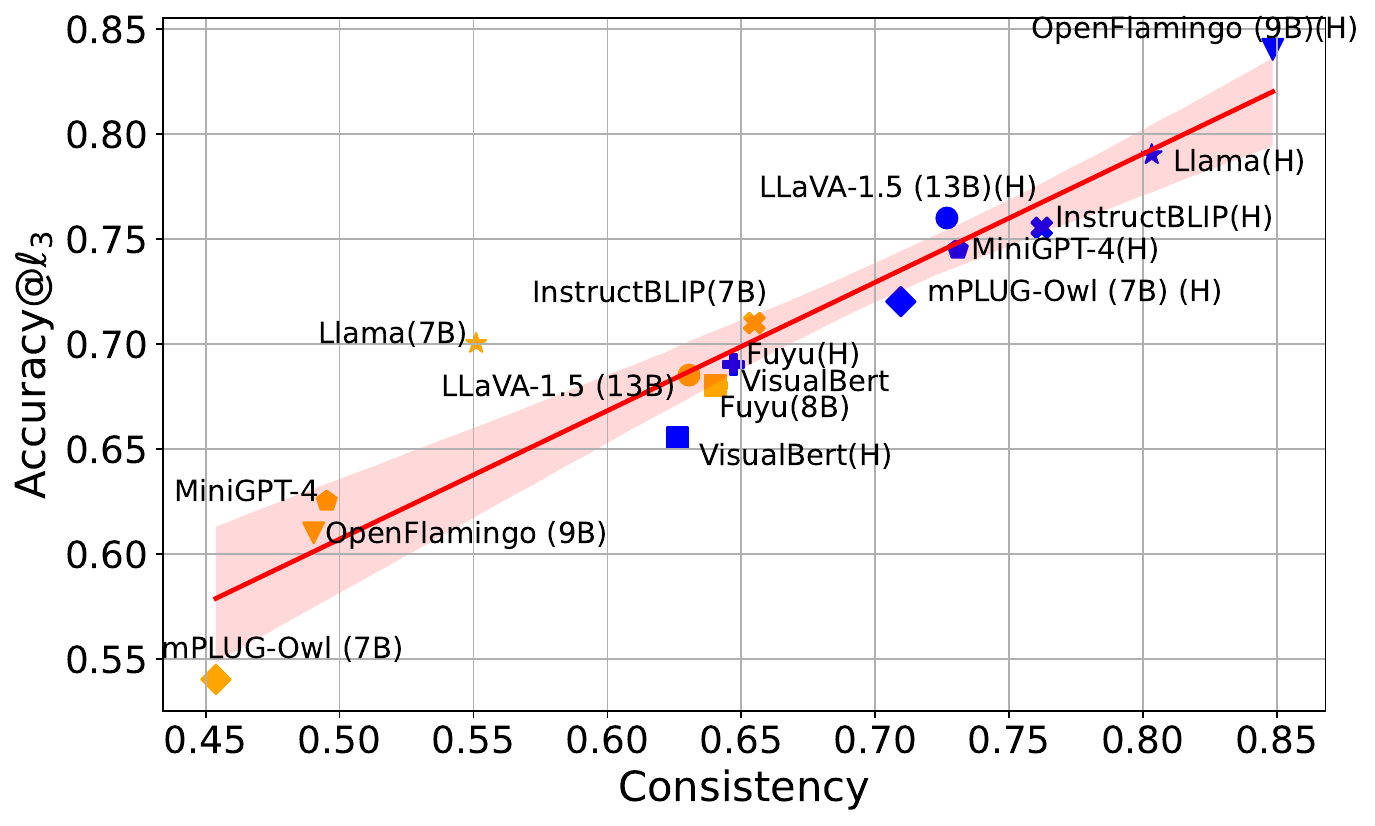}
    \caption{Relationship Between Consistency and $\ell_3$ Accuracy for Various Models with and without the TTC Layer. Orange (flat classifier), Blue (hierarchical classifier).}
   \label{fig:l2vsC}
   \vspace{-0.5cm}
\end{figure}

Overall, the hierarchical classification method generally enhances the consistency of predictions and, in many cases, improves the exact match metric. These results highlight the potential of our method to improve the performance of multimodal large language models in MLHC tasks. Figure \ref{fig:l2vsC} further supports this by demonstrating a strong positive correlation between Consistency and $\ell_3$ Accuracy, indicating that models which align their predictions with the taxonomy tend to perform better on detailed classification tasks. This correlation suggests that the TTC layer is not only effective for hierarchical classification but can also be extended to traditional classification tasks. By constructing labels from the bottom up and applying the TTC layer in a top-down manner, the divide-and-conquer approach has the potential to enhance performance in a wide range of classification problems.

\section{Conclusion}
In conclusion, the proposed taxonomy-based transitional classifier (TTC) demonstrates significant potential in enhancing the performance of large multimodal language models, particularly in hierarchical classification tasks. Across all evaluated models, the TTC layer led to notable improvements in key metrics such as Consistency, Exact Match, and $\ell_3$ Accuracy, as seen in both the grouped bar chart and the Consistency vs $\ell_3$ Accuracy diagram. While some trade-offs, such as slight reductions in HF1-Score, were observed, these were offset by the substantial gains in consistency and fine-grained accuracy, underscoring the efficacy of TTC in aligning model predictions with the underlying hierarchical structure. The strong positive correlation between Consistency and $\ell_3$ Accuracy further suggests that TTC can be extended beyond hierarchical tasks to traditional classification problems, where it could serve as a top-down, divide-and-conquer approach to boost performance. Overall, these results emphasize the versatility and effectiveness of TTC in improving both hierarchical and standard classification tasks, making it a promising addition to model-agnostic strategies for enhancing multimodal LLMs.

\section*{Limitations}
Though TTC-aided LLMs demonstrated significantly better performance across various metrics compared to traditional LLMs, and can be applied to many classification tasks, they rely on an inherent hierarchical structure in the data and require manual annotation to create multiple levels of classes. For large-scale datasets with deep hierarchies, this manual annotation incurs high labor costs, and calculating the transition matrix becomes increasingly complex. Additionally, the current approach only considers top-down transitions, ignoring bottom-up information that could enhance prediction consistency across levels. This restricts the model’s ability to capture interdependencies between lower and higher-level predictions. Moreover, the sequential nature of the TTC design limits its parallelizability, as predictions for different levels must be processed one after another. This sequential processing increases computational costs and reduces efficiency, particularly for large datasets, making the method less suited for real-time applications where speed is crucial.

\bibliography{biblio}
\begin{appendices}

 \section{The supplementary results for experiments}\label{sec:appendix}
\begin{figure}[ht]
    \centering
    \includegraphics[width=0.9985\columnwidth]{ 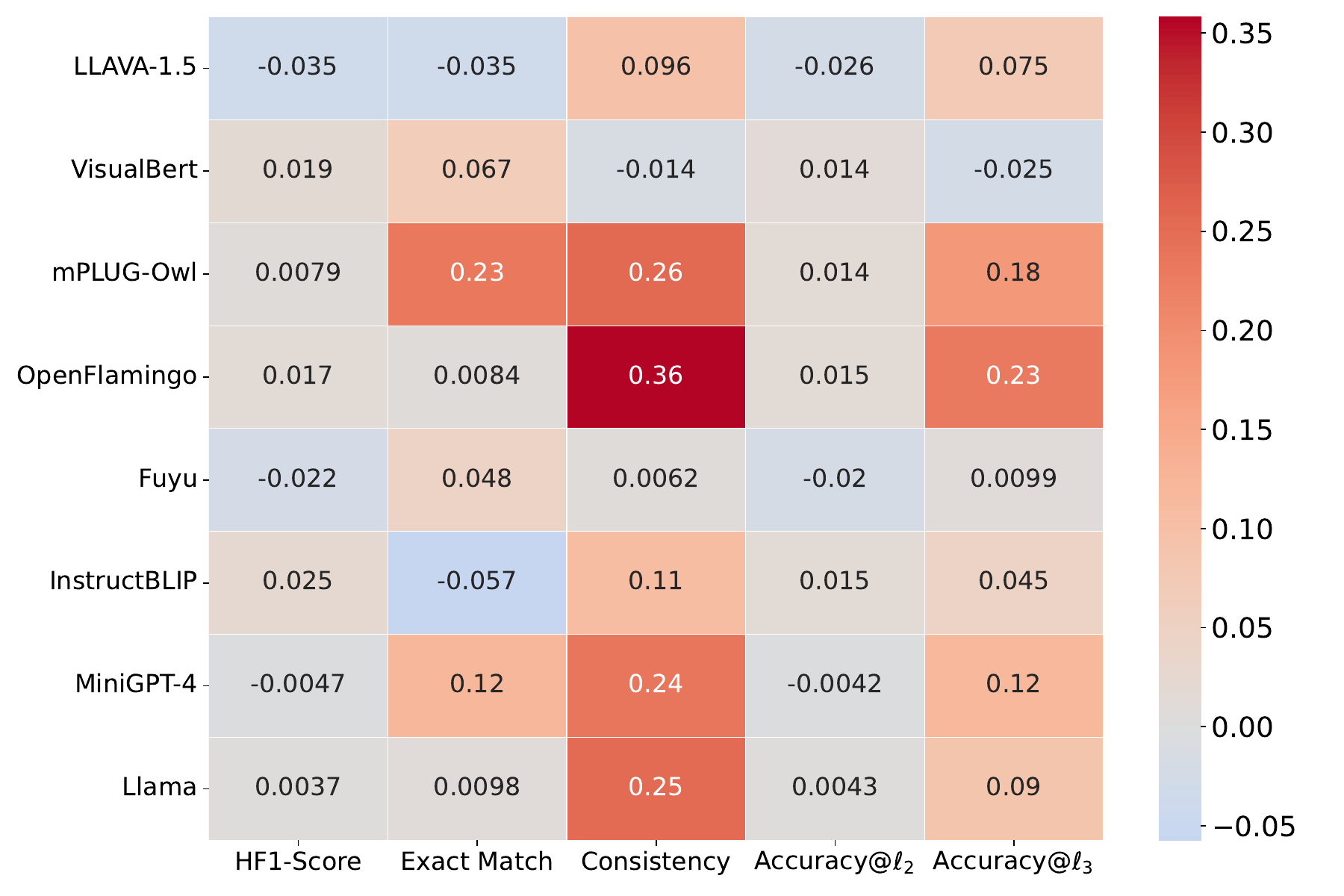}
    \caption{The heatmap of improvements that models with TTC have over models without TTC.}
    \label{fig:heatmap}
\end{figure}

The Figure \ref{fig:heatmap} heatmap illustrates the performance improvements of hierarchical (TTC) models over non-hierarchical models across five key metrics: \textbf{HF1-Score}, \textbf{Exact Match}, \textbf{Consistency}, \textbf{$\ell_2$ Accuracy}, and \textbf{$\ell_3$ Accuracy}. Positive differences, shown in red, indicate performance improvements, while blue represents declines. Notably, \textbf{OpenFlamingo (9B)} exhibits the largest improvement in \textbf{Consistency} (+0.36), while \textbf{mPLUG-Owl (7B)} shows substantial gains in both \textbf{Consistency} (+0.26) and \textbf{$\ell_3$ Accuracy} (+0.18). Although some models, such as \textbf{LLAVA-1.5 (13B)} and \textbf{InstructBLIP (7B)}, demonstrate slight decreases in \textbf{HF1-Score} and \textbf{Exact Match}, they still benefit from improved \textbf{Consistency}. Overall, the heatmap reveals that hierarchical classification consistently enhances \textbf{Consistency} and \textbf{$\ell_3$ Accuracy}, making it particularly effective for fine-grained classification tasks.

 \end{appendices}

\end{document}